\documentclass{ecai} 
\usepackage{latexsym}
\usepackage{amssymb}
\usepackage{amsmath}
\usepackage{amsthm}
\usepackage{booktabs}
\usepackage{enumitem}
\usepackage{graphicx}
\usepackage{color}
\usepackage{algorithm}
\usepackage{algorithmic}
\usepackage{tabularx}
\usepackage{float}
\usepackage{colortbl}  % 用于 \rowcolor
\usepackage[table,xcdraw]{xcolor}  % 更全面的颜色支持
\newcommand{\BibTeX}{B\kern-.05em{\sc i\kern-.025em b}\kern-.08em\TeX}

\begin{document}

%%%%%%%%%%%%%%%%%%%%%%%%%%%%%%%%%%%%%%%%%%%%%%%%%%%%%%%%%%%%%%%%%%%%%%%%

\begin{frontmatter}

%%% Use this command to specify your submission number.
%%% In doubleblind mode, it will be printed on the first page.

\paperid{123} 

%%% Use this command to specify the title of your paper.

\title{FinRobot: Generative Business Process AI Agents for \\ Enterprise Resource Planning in Finance}

%%% Use this combinations of commands to specify all authors of your 
%%% paper. Use \fnms{} and \snm{} to indicate everyone's first names 
%%% and surname. This will help the publisher with indexing the 
%%% proceedings. Please use a reasonable approximation in case your 
%%% name does not neatly split into "first names" and "surname".
%%% Specifying your ORCID digital identifier is optional. 
%%% Use the \thanks{} command to indicate one or more corresponding 
%%% authors and their email address(es). If so desired, you can specify
%%% author contributions using the \footnote{} command.

%\author{Anonymous Submission \\ For Double-Blind Review}

% \author[A]{\fnms{Hongyang}~\snm{Yang}}%\footnote{Equal contribution.}}\footnotemark
% \author[A,B]{\fnms{She}~\snm{Yang}}
% \author[A,B]{\fnms{Xinyu}~\snm{Liao}}
% \author[A]{\fnms{Likun}~\snm{Lin}}
% \author[A]{\fnms{Jiaoyang}~\snm{Wang}}
% \author[A]{\fnms{Runjia}~\snm{Zhang}}
% \author[A]{\fnms{Yuquan}~\snm{Mo}\thanks{Corresponding Author.}}
% \author[A,C]{\fnms{Christina Dan}~\snm{Wang}} 

% \address[A]{AI4Finance Foundation}
% \address[B]{Columbia University}
% \address[C]{New York University Shanghai}

% \author[A]{Hongyang Yang}
% \author[A,B]{Yang She}
% \author[A,B]{Xinyu Liao}
% \author[A]{Likun Lin}
% \author[A]{Jiaoyang Wang}
% \author[A]{Runjia Zhang}
% \author[A]{Yuquan Mo\thanks{Corresponding Author.}}
% \author[A,C]{Christina Dan Wang\footnotemark}

% \address[A]{AI4Finance Foundation}
% \address[B]{Columbia University}
% \address[C]{New York University Shanghai}

\author{
  Hongyang Yang$^{1}$,
  Likun Lin$^{1}$,
  Yang She$^{1,2}$,
  Xinyu Liao$^{1,2}$,
  Jiaoyang Wang$^{1}$,
  Runjia Zhang$^{1}$,\\
  Yuquan Mo$^{1*}$,
  Christina Dan Wang$^{1,3}$\thanks{Corresponding authors: Yuquan Mo and Christina Dan Wang. Contact: \texttt{contact@ai4finance.org}}
}
\address{
  $^1$AI4Finance Foundation\\
  $^2$Columbia University\\
  $^3$New York University Shanghai
}

\begin{abstract}
Enterprise Resource Planning (ERP) systems serve as the digital backbone of modern financial institutions, yet they continue to rely on static, rule-based workflows that limit adaptability, scalability, and intelligence. As business operations grow more complex and data-rich, conventional ERP platforms struggle to integrate structured and unstructured data in real time and to accommodate dynamic, cross-functional workflows.

In this paper, we present the first \textbf{AI-native, agent-based framework for ERP systems}, introducing a novel architecture of \textit{Generative Business Process AI Agents (GBPAs)} that bring autonomy, reasoning, and dynamic optimization to enterprise workflows. The proposed system integrates generative AI with business process modeling and multi-agent orchestration, enabling end-to-end automation of complex tasks such as budget planning, financial reporting, and wire transfer processing. Unlike traditional workflow engines, GBPAs interpret user intent, synthesize workflows in real time, and coordinate specialized sub-agents for modular task execution. We validate the framework through case studies in bank wire transfers and employee reimbursements, two representative financial workflows with distinct complexity and data modalities. Results show that GBPAs achieve up to 40\% reduction in processing time, 94\% drop in error rate, and improved regulatory compliance by enabling parallelism, risk control insertion, and semantic reasoning. These findings highlight the potential of GBPAs to bridge the gap between generative AI capabilities and enterprise-grade automation, laying the groundwork for the next generation of intelligent ERP systems.

%Through a case study on bank wire transfer workflows, we demonstrate that the system reduces end-to-end processing time by 40\%, cuts inter-step wait time by 57\%, and significantly improves overall process efficiency via action-level parallelization—while preserving full compliance and control. These results suggest that GBPAs provide a scalable and domain-adaptive foundation for intelligent ERP systems in the financial industry, effectively bridging the gap between generative AI capabilities and enterprise-grade process automation.

%Through a case study on bank wire transfer workflows, we demonstrate that the system reduces cycle time by over 50\%, lowers manual intervention by 65\%, and significantly enhances anomaly detection accuracy. These results suggest that GBPAs offer a scalable and domain-adaptive foundation for intelligent ERP systems in the financial industry, effectively bridging the gap between generative AI capabilities and enterprise-grade process automation.
\end{abstract}

\end{frontmatter}

\section{Introduction}

Enterprise Resource Planning (ERP) systems \cite{addo2011enterprise,jacobs2007enterprise,moon2007enterprise} serve as the operational backbone of financial institutions, orchestrating critical functions such as procurement, compliance, finance, and customer management. Despite their widespread adoption, traditional ERP systems remain largely dependent on static, rules-based workflows and manual configurations tightly coupled to domain-specific logic. This rigid structure limits their ability to adapt to the increasing complexity, data heterogeneity, and real-time decision-making demands of modern financial operations.

As business environments evolve, ERP systems struggle to maintain agility, cross-functional consistency, and operational responsiveness \cite{bingi1999critical,themistocleous2001erp}. Hardcoded logic and fragmented process pipelines limit real-time decision-making, delay exception handling, and increase the operational cost of change. These limitations are especially problematic in finance, where tasks such as wire transfers, risk assessments, or audits demand high accuracy, speed, and compliance.

%Fixed logic and hardcoded process definitions hinder real-time resource optimization, exception handling, and dynamic workflow adjustment in response to market shifts or client needs. The result is fragmented data pipelines, delayed decision cycles, and reduced capacity for business innovation—especially in mission-critical sectors such as finance.

%Recent advances in large language models (LLMs) and AI agent frameworks \cite{gao2023large,DeepSeek-R1,NIPS2017_3f5ee243,gpt4-report,radford2018improving,pan2024chain,huang2024understanding,wang2023survey,xi2023rise,wei2022chain} offer a paradigm shift in how enterprise systems can reason, adapt, and operate. Unlike static rule engines, LLM-based agents can interpret natural language, synthesize process knowledge, and dynamically coordinate multi-step actions across modalities. However, bringing these capabilities into ERP systems remains challenging: workflows must be auditable, agents must scale across complex tasks, and outputs must comply with strict business and regulatory requirements.

Recent advances in large language models (LLMs) and AI agent frameworks \cite{gao2023large,DeepSeek-R1,NIPS2017_3f5ee243,gpt4-report,radford2018improving,pan2024chain,huang2024understanding,wang2023survey,xi2023rise,wei2022chain} offer a paradigm shift. LLM-based agents are capable of understanding natural language, reasoning over enterprise semantics, and coordinating multi-step tasks dynamically. However, deploying LLMs within ERP systems poses unique challenges: workflows must remain traceable and auditable, agents must handle complex, domain-specific tasks, and all outputs must comply with stringent regulatory constraints.

While early efforts such as Large Process Models (LPMs) \cite{kampik2024large} and ProcessGPT \cite{beheshti2023processgpt} have explored integrating LLMs into business process management (BPM), they focus mainly on generating context-specific workflows or fragments rather than runtime orchestration or agent-based control. Similarly, general-purpose agent systems like AutoGPT, AutoGen, and Manus \cite{wu2023autogen,Significant_Gravitas_AutoGPT,Shen_manus} show potential in tool usage and autonomous reasoning but lack the structure and semantic guarantees required for ERP-grade execution. These gaps highlight the need for ERP-native agent architectures that combine LLM reasoning, business semantics, and controllable orchestration.

In this paper, we present the first AI-native, agent-based framework for ERP automation, introducing \textbf{Generative Business Process AI Agents (GBPAs)} that combine LLM reasoning, business modeling, and modular orchestration to enable adaptive workflows such as approvals, risk checks, and wire transfers. Unlike traditional engines, GBPAs dynamically interpret intent, decompose goals into executable plans, and coordinate specialized agents to fulfill them in real time. We implement the framework within a mid-sized financial institution and evaluate it on two core business processes: inter-bank wire transfers and employee reimbursements. Results show that GBPAs reduce processing time by up to 40\%, cut error rates by 94\%, and improve efficiency through intelligent parallelization—all while meeting regulatory and operational requirements. 

Our main contributions are as follows:
\vspace{-1mm}

\begin{itemize}
    \item We propose the first \textbf{AI-native, agent-based framework} for ERP systems that seamlessly integrates generative language models, domain-specific business semantics, and modular agent orchestration to enable end-to-end intelligent process automation.
    \item We adapt and extend the \textbf{Chain-of-Actions (CoA)} paradigm \cite{pan2024chain} to the ERP domain, leveraging it for dynamic task decomposition, context-aware decision routing, and multi-agent coordination across financial workflows.
    \item We implement and evaluate the framework in a production-grade banking workflow, demonstrating measurable gains in execution time, anomaly detection, and operational agility.
\end{itemize}

%%%%%%%%%%%%%%%%%%%%%%%%%%%%%%%%%%%%%%%%%%%%%%%%%%%%%%%%%%%%%%%%%%%%%%%%

\section{Related Work}

\subsection{Workflow Automation in ERP}

Workflow automation has long been a central focus in Business Process Management (BPM), with traditional ERP systems relying on rules-based engines (e.g., SAP NetWeaver, Oracle BPEL) and Robotic Process Automation (RPA) \cite{aguirre2017automation,Dalsaniya_2022,van2018robotic,mo2023bank} to encode and execute deterministic business logic. While effective in stable and repetitive environments, these systems often struggle with flexibility, particularly in handling exceptions, orchestrating cross-module workflows, and adapting quickly to evolving business requirements. As enterprise operations become more interdependent and data-driven, static automation approaches are increasingly inadequate for supporting dynamic, context-aware, and intelligent process execution.

\subsection{AI in ERP}

To address these limitations, ERP platforms have increasingly integrated AI technologies—such as machine learning, cognitive automation, and large language models (LLMs)—to improve decision support, enable intelligent process execution, and enhance cross-functional coordination \cite{Dalsaniya_2022,goundar2021artificial,haider2021artificial,wu2023bloomberggpt,ai_erp_2024,yang2023fingpt}. According to a McKinsey report \cite{ET2024}, AI-driven process optimization can increase enterprise productivity by up to 40\%. BCG further highlights that GenAI can accelerate ERP solution development by up to 5 times compared to traditional approaches \cite{BCG2025}.

Recent advances in foundation models like GPT-4 \cite{brown2020language,gpt4-report,radford2018improving}, Gemini \cite{team2023gemini}, and DeepSeek \cite{DeepSeek-R1,DeepSeek-V3} have enabled significant improvements in context understanding, multi-turn reasoning, and language-driven decision workflows across diverse data modalities.

Building on these capabilities, Kampik et al.~\cite{kampik2024large} from SAP propose \textit{Large Process Models (LPMs)}—a hybrid architecture that combines LLMs with knowledge-driven BPM to generate context-specific process variants and actionable recommendations grounded in formal process logic and enterprise performance data. Similarly, Beheshti et al.~\cite{beheshti2023processgpt} introduce \textit{ProcessGPT}, a generative AI model fine-tuned on process-centric data to assist knowledge workers with generating executable process fragments and supporting enterprise decision-making. While both works highlight the value of LLMs in BPM, they primarily focus on process-level generation rather than dynamic orchestration or modular agent-based coordination as pursued in our GBPA framework.

\subsection{AI Agents in ERP Automation}

The rise of general-purpose AI agents \cite{huang2024understanding,wang2023survey,xi2023rise,zhang2024survey,yang2024finrobot,zhou2024finrobot}, such as AutoGPT \cite{Significant_Gravitas_AutoGPT}, ReAct \cite{yao2023react}, AutoGen \cite{wu2023autogen}, Dify \cite{Charles2013}, and Manus \cite{pandayoo2024manus,Shen_manus} has demonstrated impressive capabilities in tool orchestration, autonomous reasoning, and multistep task planning. These systems show feasibility of LLM-powered agents in open-ended domains, where flexibility and breadth are prioritized. However, their underlying designs are rarely constrained by business logic, compliance rules, or domain-specific workflows, making them insufficient for high-stakes and structured enterprise environments like ERP.

To bridge this gap, several efforts have applied AI agents to enterprise scenarios. For example, agents have been used to link CRM systems with core ERP modules such as finance, inventory, and compliance \cite{Gujar2025}, enabling the execution of events-driven processes, for example, triggering invoice generation, updating stock levels, or handling fulfillment requests. Conversational agents further enhance process continuity by managing user interactions across system boundaries. More formally, Niederwieser et al.~\cite{Niederwieser2025} combine LLMs with graph-theoretical control structures and human-in-the-loop (HITL) interfaces to support ERP order workflows. While promising, these efforts typically lack modular agent pipelines, contextual process synthesis, or the ability to adaptively route tasks across execution agents.

%In contrast, our work introduces \textbf{Generative Business Process AI Agents (GBPAs)}—a domain-specific, LLM-enhanced orchestration framework for ERP environments. GBPAs dynamically interpret user intent, construct executable workflows grounded in enterprise semantics, and delegate responsibilities to modular agents specialized in document parsing, compliance validation, or financial rule enforcement. This architecture shifts ERP automation from static rule execution to reasoning-driven, real-time collaboration, supporting adaptive cognition and continuous workflow refinement.

Table~\ref{tab:erp_comparison} provides a side-by-side comparison between traditional AI-augmented ERP systems and our AI-native GBPA design, emphasizing differences in process generation, semantic control, and agent-based execution.

\begin{table}[htbp]
\centering
\caption{Comparison of AI-Augmented vs AI-Native ERP Architectures}
\label{tab:erp_comparison}
\renewcommand{\arraystretch}{1.4}
\begin{tabularx}{\linewidth}{|p{1.8cm}|p{2.5cm}|X|}
\hline
\textbf{Aspect} & \textbf{AI-Augmented ERP} & \textbf{AI-Native ERP (GBPAs)} \\
\hline
\textbf{Workflow} & Static, process definitions & Dynamic, intent-driven orchestration \\
\hline
\textbf{Interface} & Forms, scripts & Natural language, chat \\
\hline
\textbf{Data Handling} & Structured sources & Real-time, structured + unstructured \\
\hline
\textbf{Execution} & Manual triggers, rules & Agent-based automation \\
\hline
\textbf{Multi-Agent Collaboration} & Absent & Core capability \\
\hline
\end{tabularx}
\vspace{-2mm}
\end{table}

% \subsection{Process Agents vs. General-Purpose Agents}

% Process Agents (also known as Generative Business Process AI Agents) and general-purpose AI agents serve complementary roles in the design of intelligent enterprise systems. General-purpose agents are capability-focused, typically optimized for single-task reasoning, information retrieval, or execution (e.g., code generation, document parsing). They function as atomic building blocks within a larger orchestration flow.

% By contrast, Process Agents operate at the system level. They interpret end-to-end business intent, orchestrate multiple sub-agents, and dynamically generate workflows based on operational context. Process Agents are responsible for the logic-level synthesis and coordination of these general-purpose agents, forming a cognitive layer atop the functional primitives.

% This hierarchical distinction allows Process Agents to manage business semantics, enforce domain-specific constraints, and adapt workflows across systems, while leveraging general-purpose agents for execution support. This architectural separation enhances modularity, scalability, and cross-process reasoning.

%%%%%%%%%%%%%%%%%%%%%%%%%%%%%%%%%%%%%%%%%%%%%%%%%%%%%%%%%%%%%%%%%%%%%%%%

\section{Methodology}

% (EurAI, see Figure~\ref{fig:eurai}).
\begin{figure}[h]
\centering
\includegraphics[width=8cm]{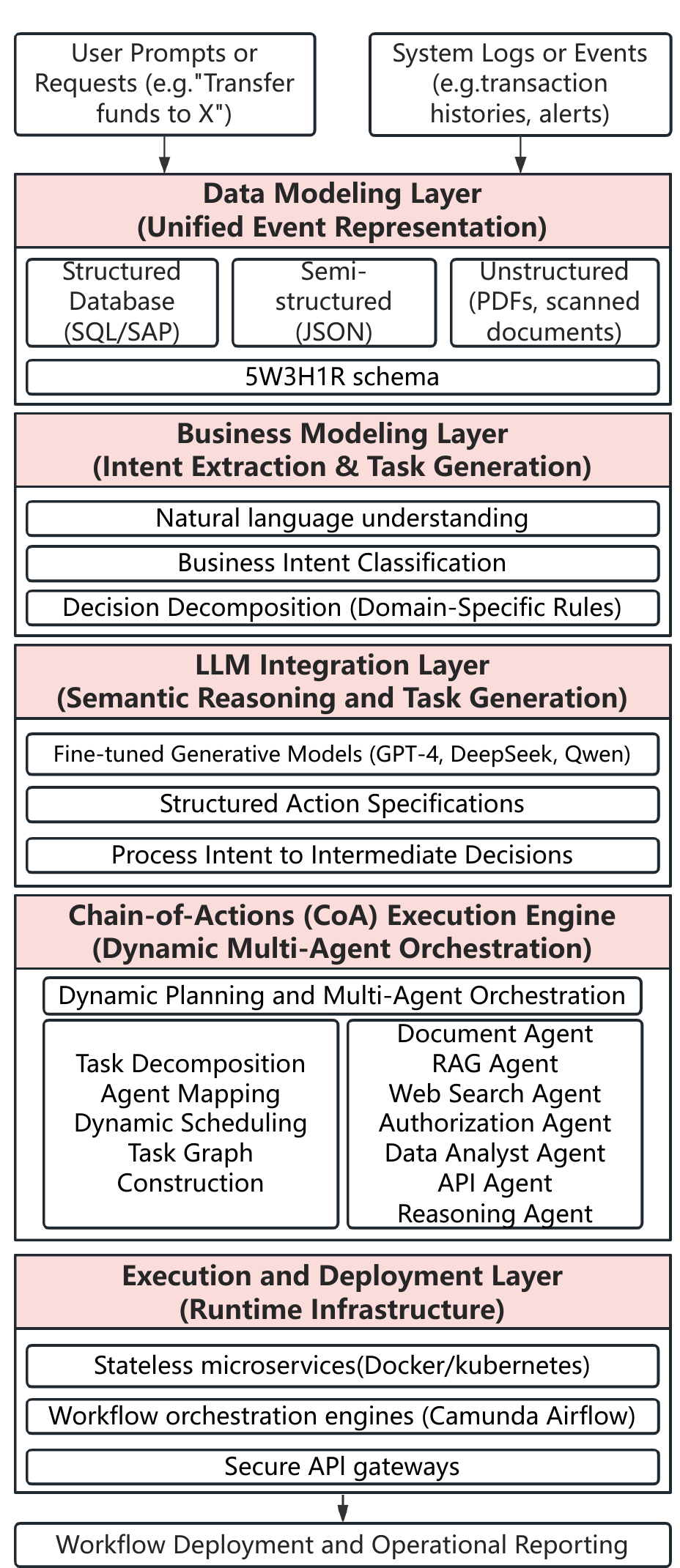}
\vspace{-2mm}
\caption{Architecture of Generarative Business Process Al Agents (GBPAs)}
\label{fig:5W3H1R}
\end{figure}

Our proposed framework for Generative Business Process AI Agents (GBPAs) follows a layered architecture that enables intelligent orchestration of enterprise workflows through modular agents, structured knowledge, and generative models. At the heart of this system lies the Chain of Actions (CoA) mechanism—a planning and execution controller responsible for dynamically composing and coordinating sub-agents in response to business intent.

\subsection{Architectural Layers}

The framework consists of five tightly integrated layers:

\begin{enumerate}[leftmargin=*]

\item \textbf{Data Modeling Layer}: This layer integrates structured (e.g., SQL, SAP), semi-structured (e.g., JSON), and unstructured (e.g., PDFs, scanned documents) data using OCR, entity recognition, and graph-based linking to create a unified, evolving enterprise knowledge base. To support LLM reasoning, we adopt an event-centric representation built on the 5W3H1R schema (Who, What, Why, When, Where, How, How much, How long, Result), capturing causal, temporal, and contextual facets of enterprise events.

\item \textbf{Business Modeling Layer}: User prompts and operational signals are translated into formal business intent. This layer parses goals, aligns them with domain-specific process templates, and enriches execution logic with contextual conditions, compliance constraints, and business semantics.

\item \textbf{LLM Integration Layer}: Fine-tuned generative models (e.g., GPT-4, DeepSeek, Qwen) serve as the reasoning core. This layer transforms process intent into intermediate decisions, retrieves task-relevant knowledge, and generates structured action specifications grounded in organizational context.

\item \textbf{Chain-of-Actions (CoA) Execution Engine}: CoA constructs an executable plan from high-level goals. Each action is mapped to a callable sub-agent—such as RAG agents, web search agents, authorization agents, data analysts, or compliance validators. CoA also manages branching, feedback loops, fallback strategies, and real-time execution scheduling. It acts as a flexible runtime planner that can respond to state changes and agent outputs.

\item \textbf{Execution and Deployment Layer}: Each agent and CoA instance is deployed as a stateless microservice within containerized environments (e.g., Kubernetes). Workflow orchestration tools like Camunda or Airflow provide scheduling, visibility, and governance. RESTful APIs, message queues, and access control layers facilitate interoperability with existing enterprise systems.
\end{enumerate}

\begin{figure}[h]
\centering
\includegraphics[width=8.7cm]{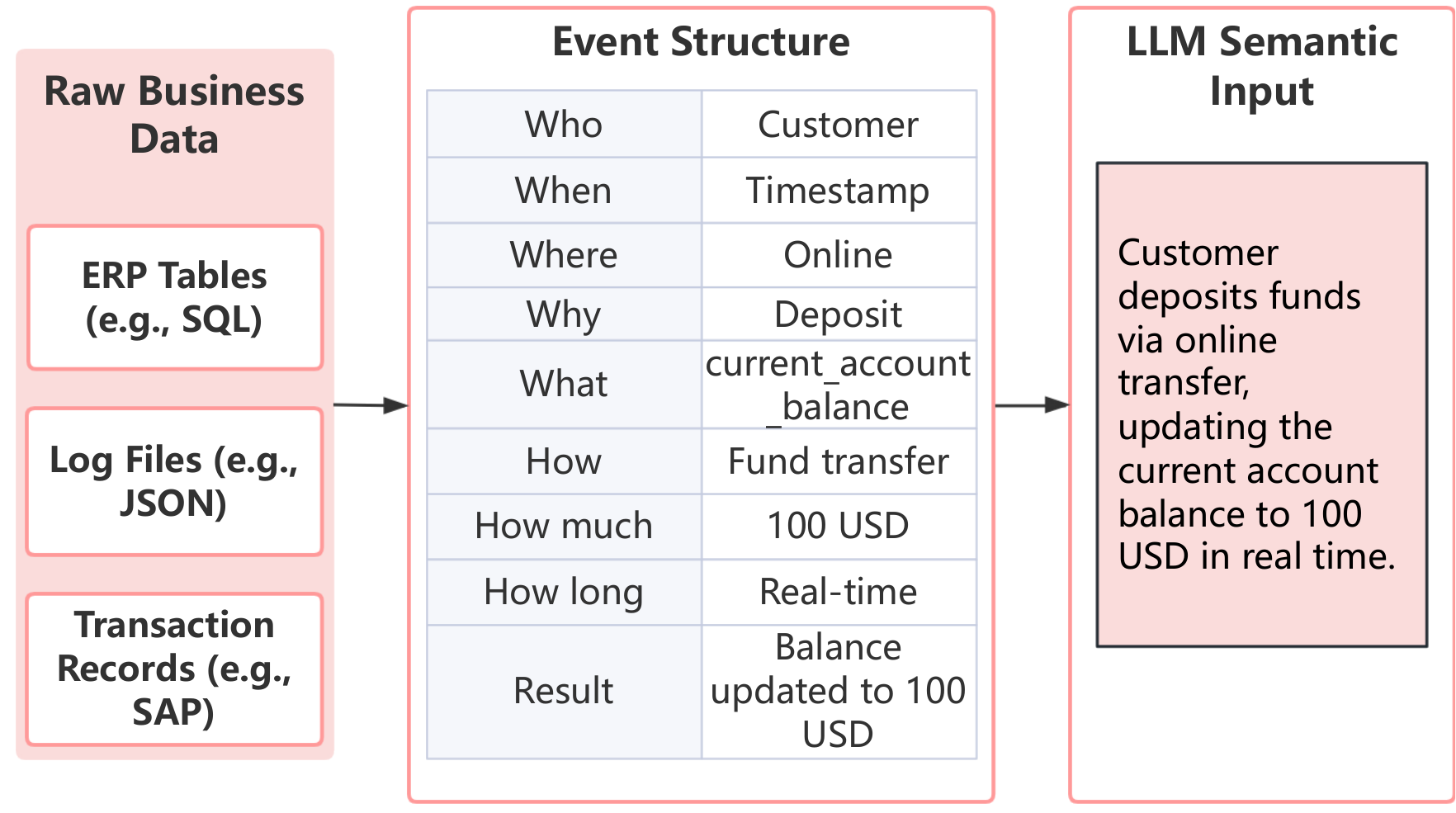}
\vspace{-2mm}
\caption{Transforming Business Data into LLM-Readable Event Semantics via 5W3H1R}
\label{fig:5W3H1R}
\vspace{-2mm}
\end{figure}

\subsection{Data Modeling Layer: Event-Centric Representation via 5W3H1R}

A major challenge in applying LLMs to enterprise systems is that business data is often fragmented and non-narrative, making it difficult for models to perform causal or contextual reasoning. To address this, we adopt a structured schema—\textbf{5W3H1R}—which reformats logs and transactions into decision-centric event representations.

The schema includes: \textit{Who, When, Where, Why, What, How, How much, How long, and Result}—capturing the full context and logic behind enterprise actions. This structure bridges the semantic gap between raw business data and LLM reasoning capabilities.

For example, a transaction such as "update account balance" in Figure~\ref{fig:5W3H1R} is encoded not just as a numerical change, but as a decision event:
the customer (\textit{Who}) performed a deposit (\textit{Why}), via a fund transfer capability (\textit{How}), which updated the CurrentAccount.Balance field (\textit{What}), in real-time (\textit{How long}), with a final result of 100 USD (\textit{Result}).

By aligning data to narrative form, the 5W3H1R schema enables LLMs to reason over enterprise workflows as coherent, causal decision sequences.

\subsection{Business Modeling Layer: Intent Extraction and Process Structuring}

The \textbf{Business Modeling Layer} translates user inputs into structured, executable process logic. It performs:

\begin{itemize}
\item \textbf{Intent Understanding}: Uses LLMs to extract business goals, entities, constraints, and expected outcomes from natural or semi-structured inputs (e.g., forms, chat).
\item \textbf{Process Template Generation}: Maps intents to parameterized workflow templates enriched with contextual rules (e.g., permissions, regulatory constraints, time conditions).
\end{itemize}

The output is a machine-readable process specification (e.g., in BPMN, JSON), passed to the CoA engine for dynamic orchestration. This layer ensures that enterprise logic is preserved while enabling modular, agent-driven execution.

\subsection{LLM Integration Layer: Semantic Reasoning and Task Generation}

The LLM Integration Layer acts as the reasoning engine of the GBPA framework. It transforms structured business intent into executable actions using fine-tuned generative models (e.g., GPT-4, DeepSeek, Qwen) tailored to enterprise tasks.

Key responsibilities include:

\begin{itemize}
\item \textbf{Fine-tuned Generative Models}: LLMs trained or adapted for domain-specific reasoning serve as the foundation for semantic interpretation and generation.
\item \textbf{Structured Action Specifications}: For each business intent, the LLM generates task-oriented instructions that define required inputs, expected outputs, constraints, and success criteria—enabling downstream agents to act autonomously.
\item \textbf{Process Intent to Intermediate Decisions}: The LLM decomposes abstract goals into executable micro-decisions, bridging high-level user prompts and fine-grained agent instructions.
\end{itemize}

Beyond deterministic rule matching, this layer supports goal-conditioned reasoning, fallback planning, and context-aware adaptation. Leveraging few-shot prompting or domain-specific fine-tuning, it ensures flexible and intelligent alignment between evolving business goals and the structured execution pipelines governed by the CoA engine.

\subsection{Chain of Actions (CoA) Execution Engine: Dynamic Multi-Agent Orchestration}

The \textbf{CoA Execution Engine} is the core execution module of the GBPA framework, translating abstract business intent into a concrete, multi-agent action plan. It dynamically decomposes user goals into task graphs and orchestrates execution using specialized agents.

Key capabilities include:
\begin{itemize}
\item \textbf{Task decomposition}: High-level business goals are broken down into executable sub-tasks.
\item \textbf{Agent mapping \& scheduling}: Each task is assigned to a suitable agent, which is dispatched in order or in parallel, depending on task dependencies.
\item \textbf{Graph-based orchestration}: An execution graph is constructed to model inter-task dependencies and control flow.
\item \textbf{Fallback \& conditional logic}: Failure cases and runtime conditions are handled with alternate branches or human-in-the-loop (HITL) escalation.
\end{itemize}

Supported agents include:
\textit{Document Agent, Retrieval Agent, RAG Agent, Web Search Agent, Authorization Agent, Data Analyst Agent, Reasoning Agent, API Agent}—each responsible for executing domain-specific tasks within the CoA pipeline.

\vspace{0.5em}
\noindent\textbf{Algorithm \ref{alg:coa}} illustrates how the CoA engine operates in practice. It parses natural language intent into a structured process plan using the LLM layer, maps each step to a corresponding agent, executes them with contextual inputs, and monitors the outcomes. Failures invoke fallback mechanisms or manual review when necessary.

\begin{algorithm}[htbp]
\caption{Chain-of-Actions (CoA) Workflow}
\label{alg:coa}
\begin{algorithmic}[1]
\STATE \textbf{Input:} Natural language intent $I$
\STATE Parse $I$ to extract user goal and constraints
\STATE Generate initial process plan $P = {s_1, s_2, \dots, s_n}$ via LLM
\FOR{each step $s_i$ in $P$}
\STATE Select agent $A_i$ from the registry
\STATE Provide context and task spec to $A_i$
\STATE Execute $A_i$ and receive output $o_i$
\IF{$o_i$ is invalid or fails}
\STATE Trigger fallback or HITL review
\ENDIF
\ENDFOR
\STATE Aggregate outputs ${o_1, \dots, o_n}$ into final result $R$
\STATE \textbf{Output:} Result $R$
\end{algorithmic}
\end{algorithm}

\noindent
This architecture decouples control flow from logic execution, enabling reusable workflows, runtime adaptability, and intelligent orchestration—key features of a truly autonomous ERP process engine.

\subsection{Execution and Deployment Layer: Scalable Agent Infrastructure}

The \textbf{Execution and Deployment Layer} provides the runtime environment for executing GBPA workflows with high scalability, modularity, and observability.

Each agent—including document, retrieval, reasoning, and API modules—is implemented as a stateless microservice, containerized via \textbf{Docker} and orchestrated using \textbf{Kubernetes}. The \textbf{Chain-of-Agents (CoA)} engine is deployed as a runtime module that dynamically executes agent workflows.

Key components include:
\begin{itemize}
\item \textbf{Workflow Engines}: Orchestration via \textbf{Camunda}, \textbf{Airflow}, or \textbf{Temporal} to manage task flow, dependencies, and agent coordination.
\item \textbf{API Gateway}: Manages secure, scalable integration with ERP/CRM/compliance systems using REST, GraphQL, or SOAP.
\item \textbf{Observability}: Uses \textbf{Prometheus} and \textbf{Grafana} for real-time monitoring, logging, and alerting.
\item \textbf{Security}: Supports scoped access control, token-based authentication, and optional service mesh policies (e.g., Istio).
\end{itemize}

This layer ensures fault-tolerant, production-ready deployment of agents, enabling dynamic and efficient execution of enterprise workflows.

Beyond runtime execution, the Execution and Deployment Layer enables direct integration of generated workflows into operational systems. Through its interface with orchestration engines and secure API gateways, GBPA-generated processes can be automatically deployed into production environments—effectively bridging intelligent planning with real-time business execution. This supports not only KPI-level monitoring and reporting, but also full-cycle delivery of workflows into ERP, CRM, and financial platforms, enabling a closed-loop system of generation, execution, and optimization.

%%%%%%%%%%%%%%%%%%%%%%%%%%%%%%%%%%%%%%%%%%%%%%%%%%%%%%%%%%%%%%%%%%%%%%%%

\begin{figure}[h]
\centering
\includegraphics[scale = 0.115]{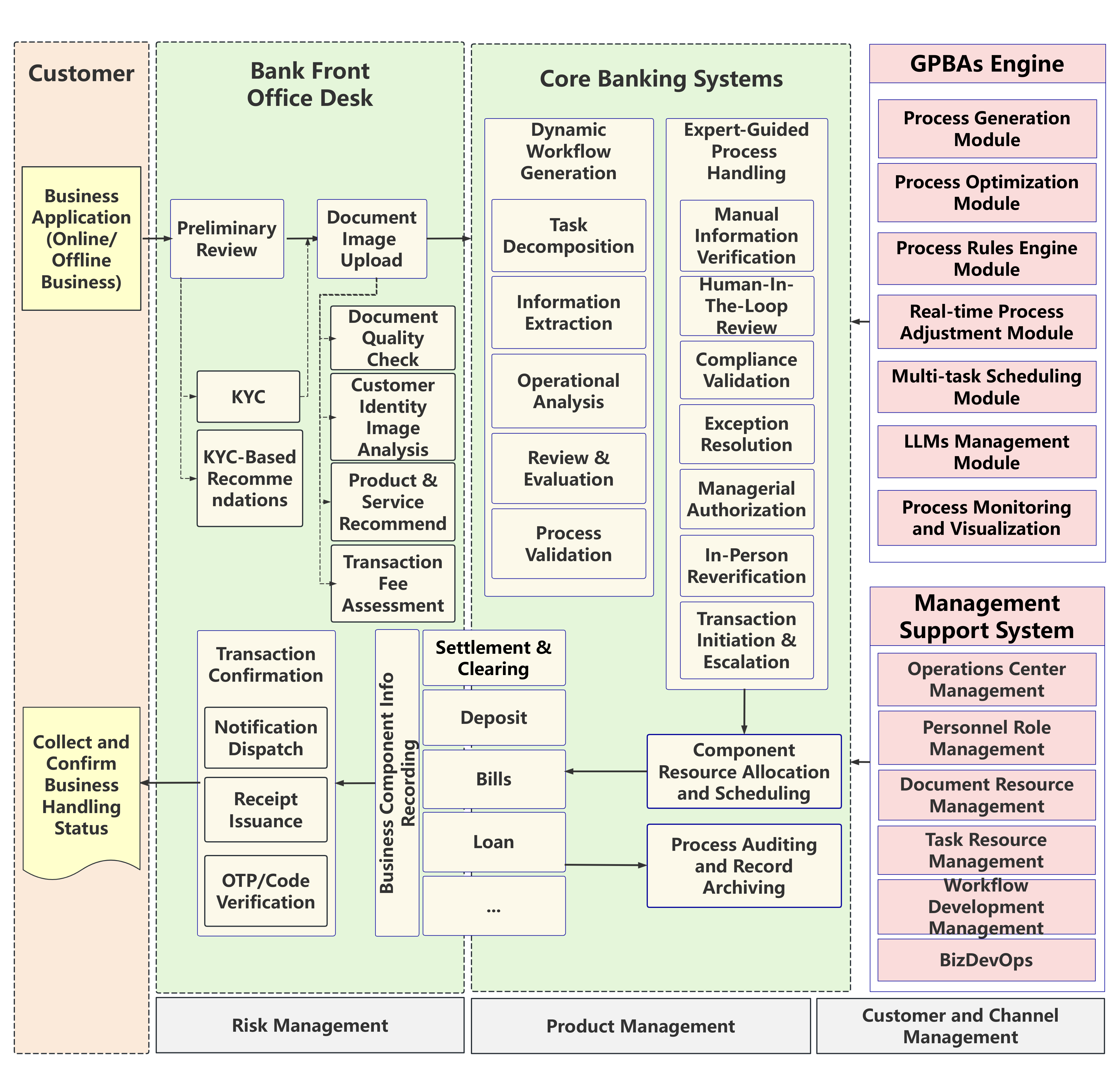}
\vspace{-2mm}
\caption{GBPAs Integrated with Core Banking Systems}
\label{fig:ai_bank}
\end{figure}

\section{Performance Evaluation}

Enterprise workflows in banking are inherently complex, often involving sequential operations such as validation, compliance, approval, execution, and post-audit processes. Traditional ERP systems rely on static BPMN diagrams or rule-based engines, which limit flexibility, scalability, and context-aware decision-making~\cite{moon2007erp, aguirre2017rpa}. In this section, we demonstrate how the proposed \textbf{Generative Business Process AI Agents (GBPAs)} framework overcomes these challenges through agent-based orchestration and LLM-driven reasoning.

\subsection{System-Level Integration: GBPAs in Core Banking Architecture}
Figure~\ref{fig:ai_bank} illustrates how GBPAs are integrated with core banking systems. The architecture connects customer-facing interfaces, bank front office operations, and backend orchestration engines into a unified digital workflow:

\begin{itemize}
\item \textbf{Front Office Desk}: Verifies submitted credentials and extracts transaction metadata through scanning, form input, and authorization procedures.
\item \textbf{Workflow Generation}: GBPAs decompose tasks and perform document layout recognition, data element extraction, and electronic seal validation.
\item \textbf{Business Processing}: Automatically populates structured data fields (e.g., account number, amount, date) and triggers business modules like voucher verification and seal authentication.
\item \textbf{Core Banking Integration}: Interacts with core modules (e.g., account systems, transaction fulfillment) while maintaining logging, monitoring, and audit functions.
\item \textbf{GPBA Engine Modules}: Enables dynamic process generation, real-time optimization, rule enforcement, LLM orchestration, and visualization.
\end{itemize}

This layered setup demonstrates GBPAs’ flexibility in transforming legacy systems into intelligent, adaptive infrastructures for end-to-end process automation.

\begin{figure}[h]
\centering
\includegraphics[scale = 0.254]{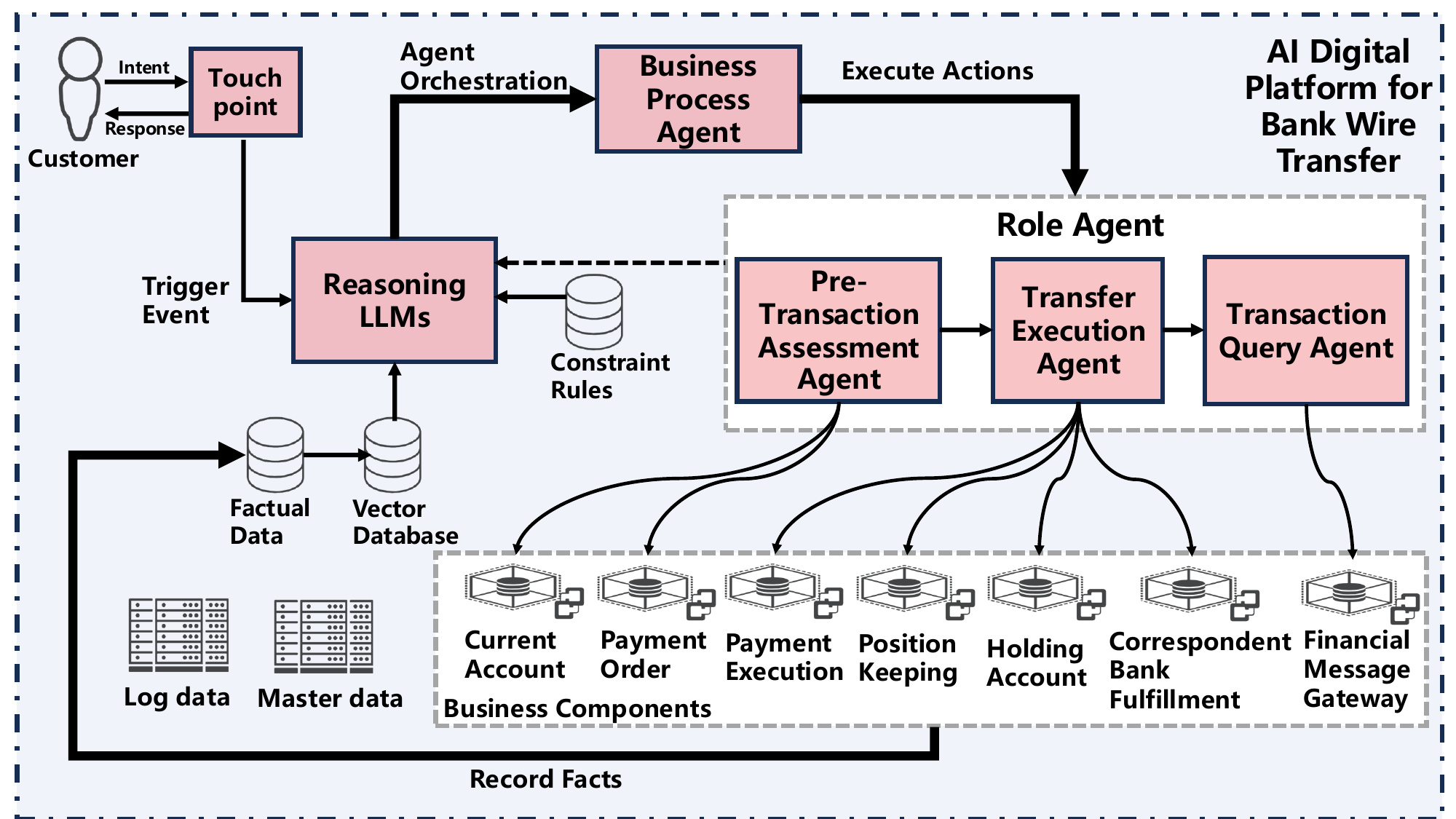}
\vspace{-2mm}
\caption{AI-Native Bank: Wire Transfer Process}
\label{fig:ai_wireflow}
\end{figure}

\subsection{Workflow Example: AI-Driven Wire Transfer Execution}

\begin{figure*}
\centering
\includegraphics[scale = 0.195]{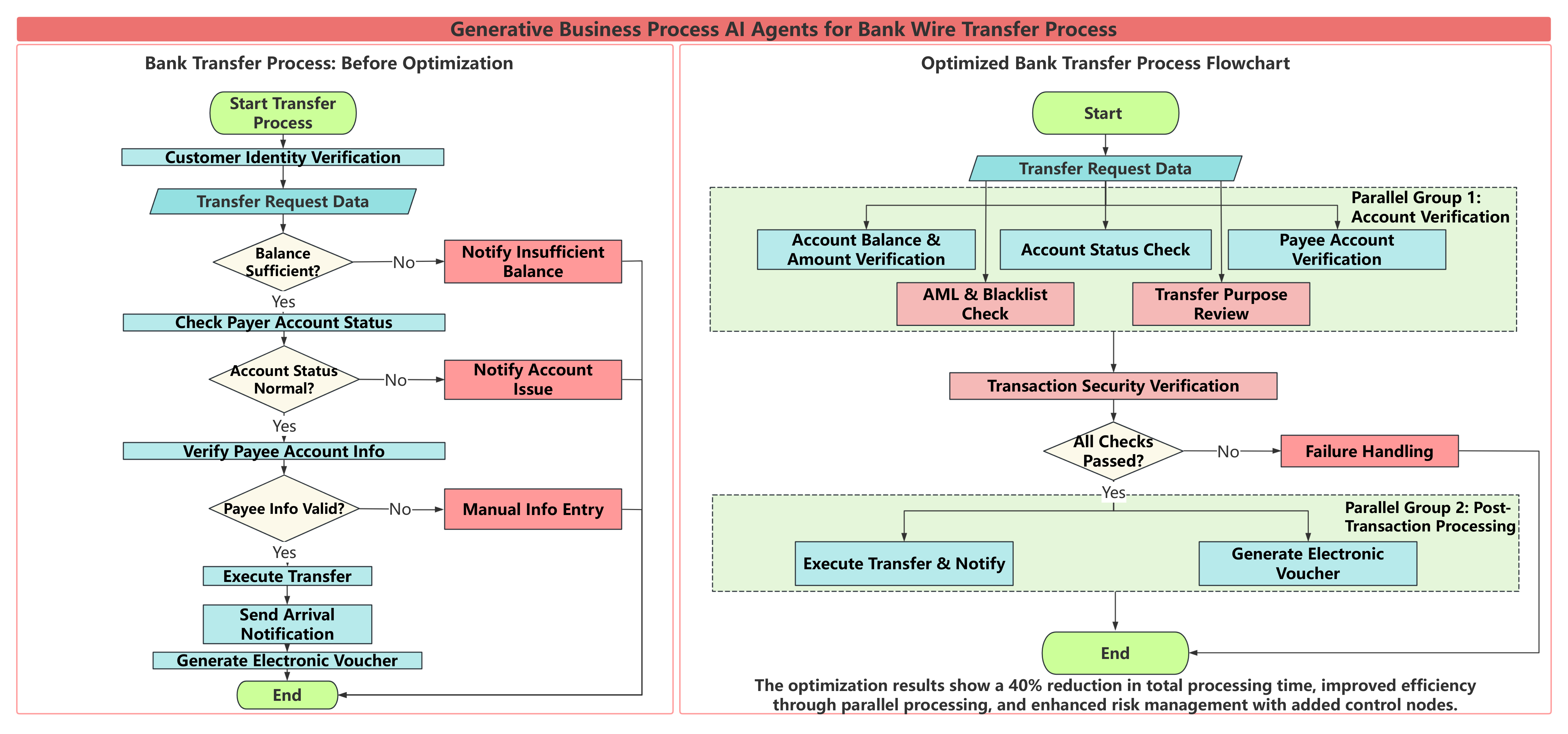}
\vspace{-2mm}
\caption{Bank Wire Transfer Process: Traditional vs. GBPAs-Optimized}
\label{fig:gbpa_banks}
\end{figure*}

As a concrete application, Figure~\ref{fig:ai_wireflow} presents the optimized wire transfer process powered by GBPAs. This workflow illustrates how customer intents trigger an LLM-based reasoning pipeline, which orchestrates a chain of specialized agents for transaction assessment, execution, and post-transaction queries.

Key components include:

\begin{itemize}
\item \textbf{Reasoning LLMs}: Interpret user intents and enforce constraint rules (e.g., AML, compliance).
\item \textbf{Role-Based Agents}: Pre-Transaction Assessment Agent, Transfer Execution Agent, and Transaction Query Agent cooperate to complete the workflow end-to-end.
\item \textbf{Modular Business Components}: Underlying services such as current account, payment execution, and correspondent bank interfaces are triggered through agent coordination.
\end{itemize}

This flow demonstrates how GBPAs dynamically orchestrate reasoning, data access, and action across traditionally siloed modules.

\subsection{Experiment Design: Comparative Evaluation Across Two Workflows}
To evaluate the generalizability of GBPAs, we apply our framework to two distinct financial workflows:

\begin{itemize}
\item \textbf{Bank Wire Transfer (Structured, Compliance-Driven)}
\item \textbf{Employee Reimbursement (Semi-Structured, Policy-Driven)}
\end{itemize}

These two use cases were intentionally selected to represent a wide spectrum of enterprise process characteristics:

\begin{itemize}
\item \textbf{Wire transfers} are mission-critical operations in the financial sector. They involve high complexity, strict compliance requirements (e.g., SWIFT integration, AML/KYC screening), and real-time execution constraints—making them ideal for validating GBPAs’ ability to handle deep, high-stakes workflows.

\item \textbf{Employee reimbursements}, by contrast, are ubiquitous across all enterprises. Though less complex, they involve high frequency, unstructured inputs (e.g., receipts, invoices), discretionary approvals, and human-in-the-loop review—making them an excellent testbed for evaluating generalizability and everyday applicability.
\end{itemize}

This dual-case evaluation ensures both depth (via complex financial workflows) and breadth (via common administrative tasks), demonstrating GBPAs’ robustness across diverse real-world ERP scenarios.

\subsection{Case Study 1: Optimizing Bank Wire Transfers with GBPAs}

\textbf{Data Description.} We first collected and modeled 607 historical transaction records using a structured 5W3H1R schema, transforming raw logs into semantically rich events suitable for LLM-driven reasoning. Using these events, the GBPA framework automatically reconstructed the end-to-end workflow, identifying 13 serially executed process nodes including balance checks, identity verification, transaction processing, and voucher generation.

\textbf{Experiment Execution.} Through multi-agent orchestration and reasoning, the GBPAs system proposed a redesigned workflow as shown in Figure~\ref{fig:gbpa_banks} with the following key improvements:
\begin{itemize}
\item \textbf{Parallelization of Independent Tasks}: Compliance checks (e.g., AML, blacklist screening), account verification, and identity validation were executed in parallel, drastically reducing idle wait times.
\item \textbf{Node Optimization}: Redundant steps were merged, and risk control stages (e.g., purpose verification) were inserted to meet regulatory demands.
\item \textbf{Execution Efficiency}: Average processing time was reduced from 15 to 9 minutes (a 40\% improvement), while inter-node wait time dropped by 57\%.
\end{itemize}

\textbf{Summary of Findings.} Table \ref{tab:gbpa_metrics_bank} shows the comparison of key metrics before and after GBPA optimization. This case highlights how GBPAs transform rigid, sequential workflows into intelligent, adaptive processes—optimizing efficiency, reducing human effort, and enhancing regulatory robustness. It demonstrates GBPAs’ potential to deliver tangible improvements in ERP-grade financial systems through agent-based orchestration and semantic process modeling.

\subsection{Case Study 2: Optimizing Reimbursement Process with GBPAs}

\begin{table*}[htbp]
\centering
\renewcommand{\arraystretch}{1.2}
\setlength{\tabcolsep}{10pt}
\begin{tabular}{|l|c|c|c|}
\hline
\textbf{Metric} & \textbf{Before GBPAs} & \textbf{After GBPAs} & \textbf{Improvement} \\
\hline
End-to-End Time & 15 min & 9 min & \textbf{-40\%} \\
Process Nodes & 13 & 9 (2 groups in parallel) & \textbf{-31\%} \\
Risk Control Stages & 0 & 2 & \textbf{+2} \\
Parallel Clusters & 0 & 2 & \textbf{+2} \\
\hline
\end{tabular}
\caption{Comparison of Key Metrics in Bank Wire Transfer Processes Before and After GBPAs Optimization}
\label{tab:gbpa_metrics_bank}
\end{table*}

\begin{table*}[htbp]
\centering
\renewcommand{\arraystretch}{1.2}
\setlength{\tabcolsep}{10pt}
\begin{tabular}{|l|c|c|c|}
\hline
\textbf{Metric} & \textbf{Before GBPAs} & \textbf{After GBPAs} & \textbf{Improvement} \\
\hline
End-to-End Time & 2.5 days & 4.25 hrs & \textbf{-82\%} \\
Process Nodes & 5 & 3 (1 group in parallel) & \textbf{-40\%} \\
Risk Control Stages & 1 & 3 & \textbf{+2} \\
Parallel Clusters & 0 & 2 & \textbf{+2} \\
Error Rate & 12.6\% & 0.8\% & \textbf{-94\%} \\
\hline
\end{tabular}
\caption{Comparison of Key Metrics in Reimbursement Processes Before and After GBPAs Optimization}
\label{tab:gbpa_metrics_reimbursement}
\end{table*}

\textbf{Data Description.} Similar to the bank wire transfer scenario, we first collected 250 historical reimbursement records and converted them into a structured 5W3H1R schema to facilitate reasoning by large language models. These records were then reconstructed and processed by the GBPA system, which automatically identified and executed 9 distinct subprocesses—including system-level validation, capital reconciliation, intelligent archiving, among others. These subprocesses were orchestrated either sequentially or in parallel, depending on contextual dependencies, to complete the end-to-end reimbursement workflow efficiently and intelligently.

\textbf{Experiment Execution.} Through multi-agent orchestration and reasoning, the system proposed a redesigned workflow with the following key improvements:

\begin{itemize}
    \item \textbf{Workflow Simplification}: Reduced from 5 sequential steps to 3 stages, including parallelized pre-check and approval, lowering node complexity by 40\%.
    \item \textbf{Faster Turnaround}: Processing time dropped from 2.5 days to 4.25 hours by enabling real-time invoice validation, budget matching, and risk flagging immediately after document upload.
    \item \textbf{Error Reduction}: Invoice error rate decreased from 12.6\% to 0.8\% through automated checks against tax systems and internal standards during submission.
    \item \textbf{Stronger Risk Control}: Expanded from 1 to 3 checkpoints, including pre-audit, smart contract-based payment, and blockchain archiving, boosting oversight by 200\%.
    \item \textbf{Parallel Execution}: Introduced 2 concurrent branches, automated pre-checks and dual approvals—cutting approval time from 24h to 4h while reducing idle wait.
\end{itemize}

%\textbf{Summary of Findings.} Table \ref{tab:gbpa_metrics_reimbursement} shows the comparison of key metrics before and after GBPA optimization. This case illustrates how process optimization based on structured event logs can significantly transform traditional reimbursement workflows. By reducing redundant steps, enabling parallel task clusters, and reinforcing risk control points, the redesigned process achieves a remarkable 82\% reduction in end-to-end time and a 94\% drop in error rate. These improvements not only streamline operations but also enhance regulatory compliance and user experience. This demonstrates the potential of data-driven, agent-enabled workflow modeling to modernize enterprise reimbursement systems with measurable efficiency and robustness gains.

\textbf{Summary of Findings.} Table~\ref{tab:gbpa_metrics_reimbursement} highlights the impact of GBPA optimization on reimbursement workflows. By streamlining steps, introducing parallel task execution, and strengthening risk controls, the revised process reduced end-to-end time by 82\% and error rate by 94\%. These gains underscore the effectiveness of agent-based, data-driven workflow modeling in improving efficiency, compliance, and user experience in enterprise reimbursement systems.

Across both cases, we observe that GBPAs:
\begin{itemize}
	 \item Drastically reduce execution time and manual workload.
	 \item Improve compliance through dynamic risk checkpoints.
	 \item Scale seamlessly across structured and semi-structured data inputs.
	 \item Enable parallelism and real-time adjustment through CoA-based orchestration.
\end{itemize}
These results highlight the transformative impact of integrating LLM-driven agent frameworks in enterprise financial systems. By converting procedural logic into adaptive, interpretable, and composable workflows, GBPAs establish a scalable foundation for next-generation ERP automation. 

\section{Conclusion}

This paper presents a novel architecture for Generative Business Process AI Agents (GBPAs), an AI-native framework that integrates LLM reasoning, modular agent orchestration, and structured business modeling for ERP automation. GBPAs enable dynamic, interpretable workflows through CoA-based planning and a unified 5W3H1R schema.

Evaluated on real-world financial scenarios, GBPAs significantly improve efficiency, accuracy, and compliance. As ERP systems evolve, our framework provides a scalable foundation for AI-driven enterprise transformation.

\section{Discussion and Future Outlook: From AI-Augmented Tools to AI-Native Enterprises}

\subsection{From Fragmented Tools to AI-Native Architectures}

Most enterprises still treat AI as a modular enhancement—applied to isolated workflows such as marketing automation, customer service, or financial analytics. This fragmented approach positions AI merely as a “tool for people,” reinforcing the traditional structure of functionally segmented, human-centric workflows.

However, we argue that this perspective underestimates the transformative potential of AI. The essence of a modern enterprise is the coordination of specialized human functions through systems, rules, and data flows. These divisions, while necessary under human constraints, introduce substantial entropy—misunderstandings, handoffs, misaligned data, duplicated effort, and interdepartmental inefficiencies.

AI-native enterprises offer a structural alternative. Instead of embedding AI within legacy silos, these organizations treat data as the primary asset and intelligent agents as the computational fabric that dynamically composes and executes workflows. Departments are replaced by autonomous microservices; workflows are driven by generative agents; and human roles shift from task execution to strategic oversight. This marks a paradigm shift from process automation to organizational cognition.

\subsection{AI Literacy: A New Organizational Capability}

Yet realizing this shift requires more than infrastructure—it demands organizational readiness. A key enabler of this transition is \textbf{AI literacy}, which encompasses the competencies required to evaluate, communicate, and collaborate effectively with AI systems. At the individual level, AI literacy includes understanding how models make decisions, interpreting AI outputs, and engaging in human-AI collaboration. At the organizational level, it becomes a dynamic capability that reinforces strategic agility and innovation~\cite{cetindamar2022explicating, long2020ai, helfat2003dynamic}.

This need is no longer optional. Under the EU AI Act, which enters into force on February 2, 2025, AI system providers and operators are required to ensure sufficient levels of AI literacy among personnel responsible for the deployment and operation of AI tools. Article 4 mandates this competency as a prerequisite for trustworthy AI deployment in regulated environments.

The proposed GBPA framework supports this need by making AI systems \emph{interpretable, auditable, and actionable}. Its agent-based design allows employees to directly observe how their decisions influence process outcomes, turning AI literacy from an abstract goal into a tangible organizational KPI. 

\subsection{Financial Services: A Natural Domain for AI-Native Orchestration}

This transformation is particularly well-aligned with the \textbf{financial services industry}, where core functions—risk assessment, transaction processing, compliance, and advisory—are intrinsically information-driven. Financial institutions function as structured pipelines of rules, records, and decisions, making them ideal candidates for AI-native orchestration. Intelligent agents can plan, reason, and execute financial workflows end to end with traceable logic and real-time adaptability.

In sum, GBPAs offer a scalable path to AI-native enterprises by unifying LLM reasoning, agent orchestration, and interpretable workflow generation—particularly suited for data-driven, compliance-intensive domains like finance.

%%%%%%%%%%%%%%%%%%%%%%%%%%%%%%%%%%%%%%%%%%%%%%%%%%%%%%%%%%%%%%%%%%%%%%%%

%\bibliography{citation}

\bibliography{citation}

\end{document}